\documentclass[runningheads]{llncs}

\usepackage{eccv}

\usepackage{eccvabbrv}

\usepackage{graphicx}
\usepackage{booktabs}

\usepackage[accsupp]{axessibility}  

\usepackage[pagebackref,breaklinks,colorlinks,citecolor=eccvblue]{hyperref}

\usepackage{orcidlink}

\begin{document}

\title{Towards motion from video diffusion models} 


\author{Paul Janson\inst{1,2} \and
Tiberiu Popa\inst{1} \and
Eugene Belilovsky\inst{1,2}}

\authorrunning{P.~Janson et al.}

\institute{Concordia University, Montreal , Canada \and
MILA Quebec AI Institute, Montreal, Canada \\
\email{p\_janso@live.concordia.ca}}

\maketitle

\begin{abstract}
Text-conditioned video diffusion models have emerged as a powerful tool in the realm of video generation and editing.  
But their ability to capture the nuances of human movement remains under-explored. 
Indeed the ability of these models to faithfully model an array of text prompts can lead to a wide host of applications in human and character animation. 
In this work, we take initial steps to investigate whether these models can effectively guide the synthesis of realistic human body animations. 
Specifically we propose to synthesize human motion by deforming an SMPL-X body representation guided by Score distillation sampling (SDS) calculated using a video diffusion model.
By analyzing the fidelity of the resulting animations, we gain insights into the extent to which we can obtain motion using publicly available text-to-video diffusion models using SDS.
Our findings shed light on the potential and limitations of these models for generating diverse and plausible human motions, paving the way for further research in this exciting area.
  \vspace{-3pt}
\end{abstract}
\section{Introduction}
\label{sec:intro}
Video generative models\cite{imagen_video,villegas2022phenaki,sora,modelscope,videocrafter2,make_a_video} have been shown and claimed to be potential tools for simulating the world. Recent advancements, such as those demonstrated by \cite{sora}, highlight the capabilities of video diffusion models trained on vast datasets to generate realistic and diverse visual content. These impressive capabilities inspired us to question the potential of open-source counterparts for the specific task of generating human motion animations from natural language input. 
We sought to determine if the current state-of-the-art open-source models can be used with score distillation to generate human motion effectively.

Human motion generation from textual instructions is a well-studied domain\cite{action_gul_varol,mdm,avatarone,motionclip}, with recent methods often involving diffusion models trained on motion capture (MoCap) data\cite{mdm,priormdm}. However, the scarcity of MoCap data compared to the abundance of video data presents a scalability challenge. Extracting animation directly from videos, mirroring the success seen in 3D asset generation\cite{get3d,clipmesh,magic3d,prolificdreamer}, represents an ambitious yet necessary goal. 


We define our problem as determining the correct sequence of joint rotations necessary to produce realistic human motion described by a prompt. To achieve this, we utilize the widely adopted SMPL-X\cite{smplx} digital human template model to render a character. This character is animated through an optimization process that iteratively updates a multi-layer perceptron (MLP) to predict the corresponding pose parameters. We guide this optimization process using a video diffusion model, which provides feedback on the realism of the generated motion.

\begin{figure}[!t]
    \centering
    \includegraphics[width=\linewidth]{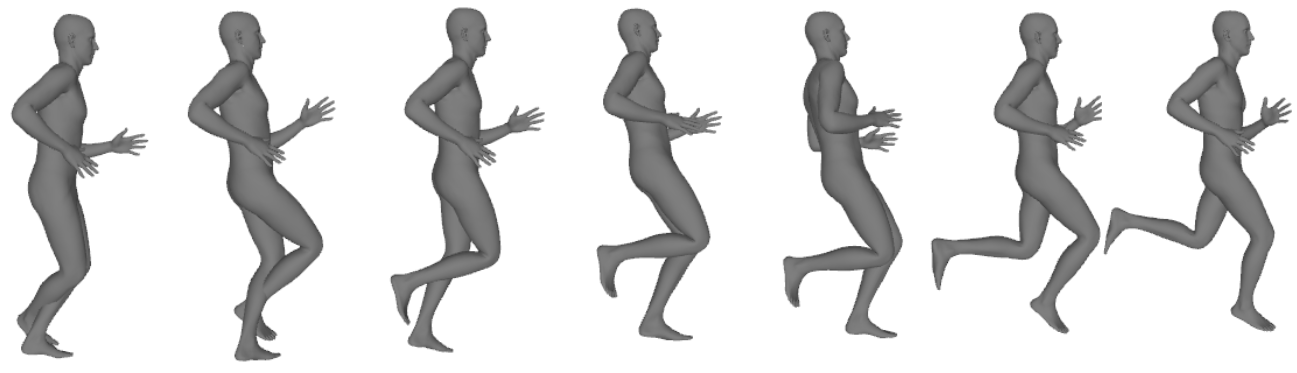}
    \caption{Human motion sequence resembling running generated using text-to-video model. The figure illustrates that the current video models can generate realistic motion for commonly occuring human activity such as running}
    \label{fig:teaser}
     \vspace{-15pt}
\end{figure}

As shown in Figure \ref{fig:teaser}, these models demonstrate a proficiency in generating animations for common actions such as running. However, when tasked with depicting uncommon or rare human movements, their performance under score distillation reveals limitations.  While they can produce visually appealing results for familiar actions, their ability to capture the nuances of less frequent movements remains a challenge.
Our main contributions are as follows:
\vspace{-2pt}
\begin{itemize}
    \item We propose a differentiable video generation pipeline: \textbf{MotionDistill} that leverages text-to-video diffusion models to generate human motion.
    \item We conduct an evaluation of ModelScope\cite{modelscope}, ZeroScope\cite{zeroscope}, and VideoCrafter\cite{videocrafter1,videocrafter2}'s capabilities in generating both common and uncommon human motion.
    \item We ablate our analysis to the latent space of these models to test the effectiveness of SDS
\end{itemize}
\vspace{-8mm}
\section{Related Works}
\vspace{-2pt}
\textbf{Diffusion models in content creation:}
Recent advances in text-to-image\cite{imagen,stablediffusion} foundation models \cite{foundationmodels} have catalyzed exploration into their application for three-dimensional (3D) asset synthesis\cite{sivakumar2024fewshotnerf} and editing\cite{zamani2024temporally}. Early works, particularly with CLIP\cite{clip}-based models , leveraged joint image-text embeddings to generate 3D assets directly from text prompts \cite{clipmesh,avatarone,dreamfields,clip_forge,clip_nerf,dream3d}. However, the emergence of diffusion models \cite{diffusion_beats_gan,diffusion_model} and the introduction of SDS \cite{dreamfusion} marked an important shift. SDS enabled the extraction of 3D assets\cite{prolificdreamer,dreamfusion,dreamgaussian,magic3d,get3d} as NeRFs\cite{nerf}, Meshes\cite{dmtet,tada,dreamavatar} and Gaussian splats\cite{gaussian_splats}.\cite{make_a_video_4d} integrated a temporal dimension, facilitating the generation of animated 2D\cite{breathinglife} and 3D assets from video diffusion models.\cite{4dfy,dreamin4d,align_your_gaussians}, introduced hybrid SDS methodologies and alternative representations like Gaussian splattings \cite{align_your_gaussians} to enhance the fidelity and motion quality of generated assets. However, these approaches focused on open-ended generation and did not specifically address the potential of video models to generate diverse human motions.\\
\textbf{Text to Human motion generation:}
The generation of human motion guided by textual descriptions is a well-established area of research. Initial approaches explored to model it as machine translation \cite{learning_bi_mapping,text2action} and joint cross-modal mappings \cite{ghosh2021synthesis,language2pose} to address this challenge. Subsequent works leveraged motion capture datasets\cite{kitmotionlanguage,amass} to train models capable of generating human motion as sequences of poses. Variational Autoencoders (VAEs)\cite{vae} were employed in \cite{petrovich2022temos, guo2022generating}. The concept of a shared latent space with CLIP\cite{clip} was introduced in \cite{motionclip}. Human motion diffusion model \cite{mdm} pioneered the application of diffusion-based modeling to human motion generation, enabling ancestral sampling in the motion space, which subsequently led to the utilization of SDS\cite{dreamfusion} in \cite{priormdm} for generating extended motion sequences.
These prior methods are limited by motion capture data, operating solely in the rotation space. Our approach ventures into the pixel space models, exploring their potential in motion generation.
\vspace{-5mm}
\section{Background}
\textbf{Human template models - SMPLx:}
\label{sec:smplx}
The Skinned Multi-person Linear(SMPL)\cite{smpl,smplx,mano,flame} family of models comprises articulated human body models parameterized by the shape and pose parameters. Among these, We use SMPLx because of its comprehensive representation of the human body. 
It is defined by the function $M(\theta,\beta,\phi): \mathbb{R}^{|\theta| \times |\beta| \times |\psi|} \rightarrow \mathbb{R}^{3N}$ \cite{smplx} given by Equation \ref{eqn:smplx}.
\begin{equation}
 \label{eqn:smplx}
    M(\beta,\theta,\psi) = LBS(T_p(\beta,\theta,\psi),J(\beta),\theta,\mathcal{W})
\end{equation}
This function takes body parameters as input, performs linear blend skinning, and outputs a mesh with $N=10,475$ vertices. Here $\theta \in \mathbb{R}^{3(K+1)}$ denotes the body pose parameters, $\beta \in \mathbb{R}^{|\beta|}$ denotes the  shape parameters and $\psi \in \mathbb{R}^{|\psi|}$ denotes the facial expression parameters. $\mathcal{W}$ is the skinning weights and $J$ is the joint regressor. The pose parameters $\theta$ can be further divided into $\theta_b$ (body joints pose), $\theta_f$ (jaw pose) and $\theta_h$ (finger pose). While SMPLx accounts for $K=54$ body joints, our approach optimizes only the major body joints $K_{b}=21$ focusing exclusively on $\theta_b$ to constraint our scope. The template body $T_{p}(\beta,\theta,\psi)$ is given by Equation \ref{eqn:smplx_template}. Here $\hat{T}$ is the template mesh, $B_S, B_E, B_P$ denotes the blend shape functions corresponding to shape, expression, and pose. 
\begin{equation}
    \label{eqn:smplx_template}
    T_{p}(\beta,\theta,\psi) = \Bar{T} + B_{S}(\beta;\mathcal{S}) + B_{E}(\psi;\mathcal{\mathcal{E}}) + B_{P}(\theta;\mathcal{P})
\end{equation}
\textbf{Score Distillation Sampling:}
\label{sec:sds}
Score Distillation Sampling (SDS)\cite{dreamfusion} is a method employed to leverage large-scale diffusion models\cite{ddpm,diffusion_model} for training compact parametric image generators. It utilizes the score function\cite{scorebasedgen} of the diffusion model to derive gradient directions for updating the generator, iteratively aligning it with provided textual prompts. 
Our formulation of SDS slightly deviates from the original by employing a latent diffusion model\cite{ldm}.Given a pre-trained latent diffusion model $\phi$ with its denoising UNet $\hat{\epsilon}_{\phi}(z_t;y,t)$ \cite{unet}, text prompt $y$ and an image generator parameterized by $\theta$. The gradients needed to update the generator are obtained by Equation \ref{eqn:sds}. Here $t$ is the randomly sampled diffusion timestep, and $\epsilon$ is the randomly generated noise. $z$ is the generated image encoded to the latent space of the diffusion model and $z_t$ is the noised image and $w(t)$ is a weighting function.  \cite{ldm,diffusion_model}.  
\begin{equation}
    \label{eqn:sds}
    \nabla_{\theta}\mathcal{L}_{SDS}(\phi,z) = E_{t,\epsilon}[w(t)(\hat{\epsilon}_{\phi}(z_t;y,t)-\epsilon) \frac{\partial z}{\partial \theta}] 
\end{equation}
\begin{figure}[t]
    \centering
    \includegraphics[width=1.0\textwidth]{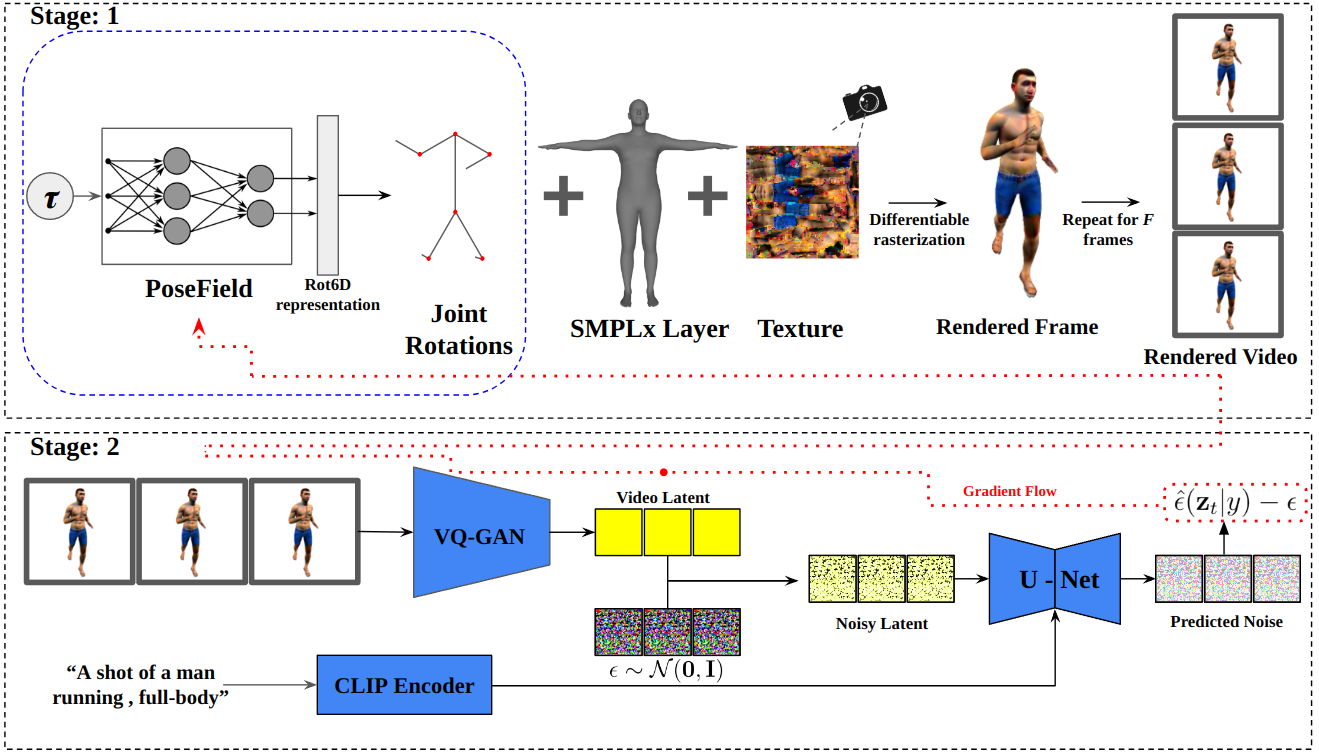}
    \caption{Our study consists of two stages. Stage:1 (top) Joint rotations required to animate the character are generated using PoseField. Passed through SMPLx Layer to get the final mesh which is then rasterized using a differentiable renderer. We use a random camera and a predetermined texture. This is repeated for F frames to obtain the video. Stage:2 (bottom) Rendered video is encoded to the latent space of the diffusion model then random noise is added to the latent. Unet of the Video model is used to predict the added noise and the gradients are estimated by SDS.}
    \label{fig:method}
\end{figure}
\vspace{-15mm}
\section{Method}
In this section, we detail our proposed methodology.
The central challenge we address is as follows: 

Given a text prompt $\textbf{y}$ describing an action, we aim to determine the optimal joint rotations $\theta_{b}$ for each frame $\{\theta_{b_{1}},\theta_{b_{2}},... \}_{F}$ of an $F$-framed video that aligns with the given prompt. 
 To tackle this, we model the rotations implicitly using a neural network "PoseField" $P_{\alpha}:\tau \rightarrow \theta_b$ parameterized by $\alpha$. It is a Multi-Layer Perceptron (MLP) comprising two hidden layers. This network takes the frame id $\tau$ as input and outputs the corresponding body pose parameter $\theta_b$.  Consequently, our task shifts from finding the optimal rotation set to determining the optimal parameters $\alpha^{*}$ of the network $P_\alpha$.
 
\noindent As illustrated in Figure \ref{fig:method}, our approach employs a two-stage process.
\begin{enumerate}
    \item \textbf{Video generation:} We utilize a differentiable pipeline to synthesize a video of the animation sequence.
    \item \textbf{Gradient Estimation:} We leverage video diffusion models to estimate gradients for updating the PoseField parameters.
\end{enumerate}
 \subsection{Stage 1: Generating multi-view video of the animation sequence}
In this stage, we generate the video of the animation sequence frame by frame and subsequently concatenate these frames to form a complete video.
We encode the frame identifier $\tau$ with positional encoding\cite{transformer} and infer the PoseField $P_{\alpha}$ to obtain the SMPL-X body pose parameter $\boldsymbol{\theta}_{b} = P_{\alpha}(\tau)$ corresponding to the current frame. Using this, we derive the mesh $M(\boldsymbol{\beta}, [P_{\alpha}(\tau), \boldsymbol{\theta}_f, \boldsymbol{\theta}_h], \boldsymbol{\psi})$. It is important to note that only $\alpha$ is the trainable parameter here, while all other values are reused across frames and training iterations.
We render the character from a randomly selected camera position sampled from a circular trajectory around the mesh. Given the camera trajectory $C$ and the rendering function $R$, we obtain the projection matrix $\pi$ and render the current frame as $I_{\alpha} = R(\pi(M))$. This process is repeated for $F$ frames, generating a series of frames ${I_1, I_2, \ldots, I_F}$. We concatenate these frames to obtain the video $V_{\alpha}$ from a certain view that will be used as input to the next stage for gradient estimation.
\vspace{-5mm}
\subsection{Stage 2: Estimating gradients for update}
The generated video is then used to estimate the gradient using Score Distillation Sampling (SDS)\cite{dreamfusion}. We employ the temporal variation of SDS as proposed in \cite{make_a_video_4d}. Since we use Latent Diffusion Models, we first encode the video to the latent space of the video diffusion model: $Z_{\alpha} = E(V_{\alpha})$. We then add noise to the video latent according to the noise schedule: $Z_{(\alpha,\sigma,\epsilon)} = \sqrt{1-\sigma^{2}}Z_\alpha + \sigma\epsilon$, where $\epsilon$ is randomly generated noise and $\sigma \in (0,1)$ is the noise level.
Gradients are then computed by Equation \ref{eqn:sds_t}. Here $\lambda_{SDS_t}$ is a hyperparameter that controls the effect of SDS.
\begin{equation}
    \label{eqn:sds_t}
    \nabla_{\theta}\mathcal{L}_{SDS-T} = \lambda_{SDS_t} E_{\sigma,\epsilon}[w(\sigma) \hat{\epsilon}(Z_{\alpha,\sigma,\epsilon}|y,\sigma) - \epsilon) \frac{\partial Z_{\alpha}}{\partial \alpha}]
\end{equation}
\textbf{Regularization:}
To encourage smoothness in the generated motion, we introduce a regularization constraint. We add the following loss term, which minimizes the difference between the body poses of consecutive frames. Here $\lambda_{reg}$ is a regularization coefficient.
\begin{equation}
    \mathcal{L}_{reg} = \lambda_{reg} \sum_{i=1}^{F-1} (\theta_{b_{i+1}} - \theta_{b_{i}})
\end{equation}
\textbf{SDS from Stable diffusion model:}
Following recent works\cite{align_your_gaussians,4dfy,make_a_video_4d} and considering the fact that video diffusion models often lack visual quality compared to their image counterparts, we additionally estimate gradients using the standard SDS with an Image Diffusion Model. We treat each video as a batch of images and calculate the gradients using Equation \ref{eqn:sds}.
\section{Experiments and results}
\vspace{-4mm}
\begin{figure}[h!]
    \centering
    \includegraphics[width=0.9\textwidth]{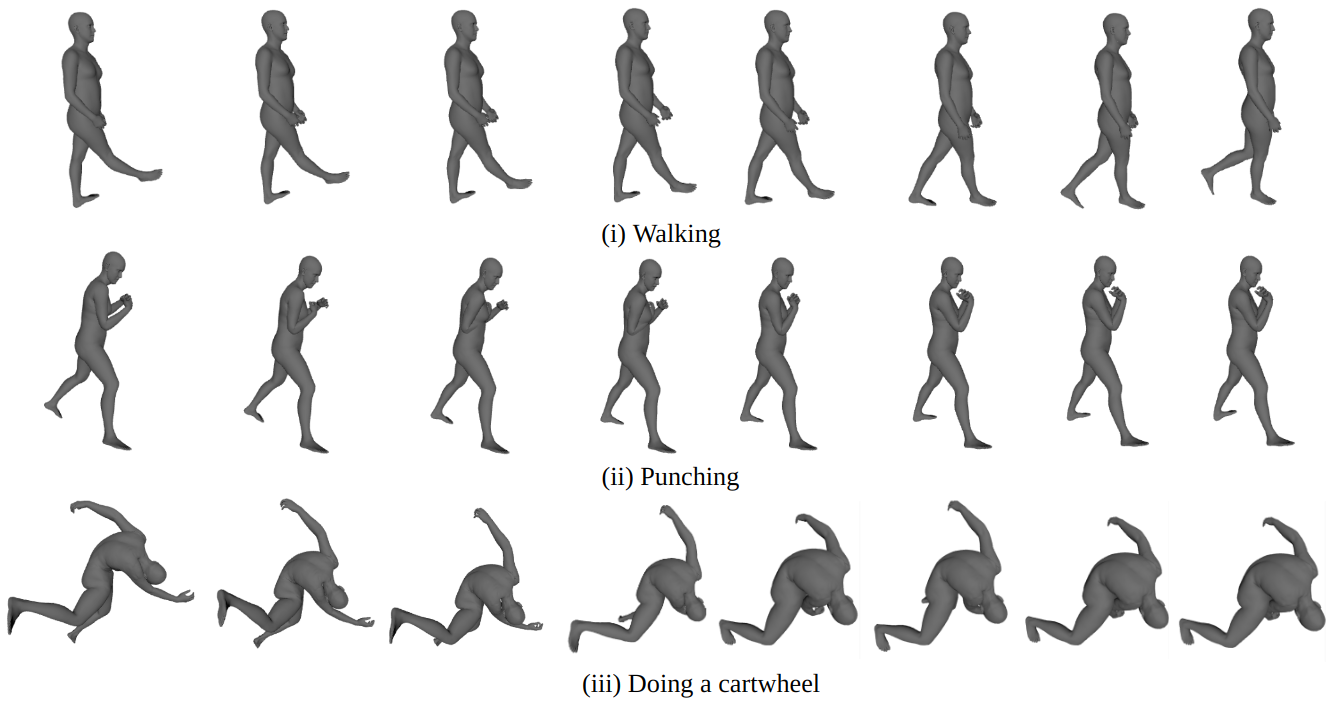}
    \caption{\textbf{Our results for different motions:} All results are obtained by using the model VideoCrafter\cite{videocrafter2}. (i) Walking motion is one of the best cases in addition to running (Fig:\ref{fig:teaser}). (ii) Punching is a semi-failure case (iii) Cartwheel is an extreme failure case. }
    \label{fig:results}
    \vspace{-8mm}
\end{figure}

\begin{figure}[t]
    \centering\includegraphics[width=0.95\textwidth]{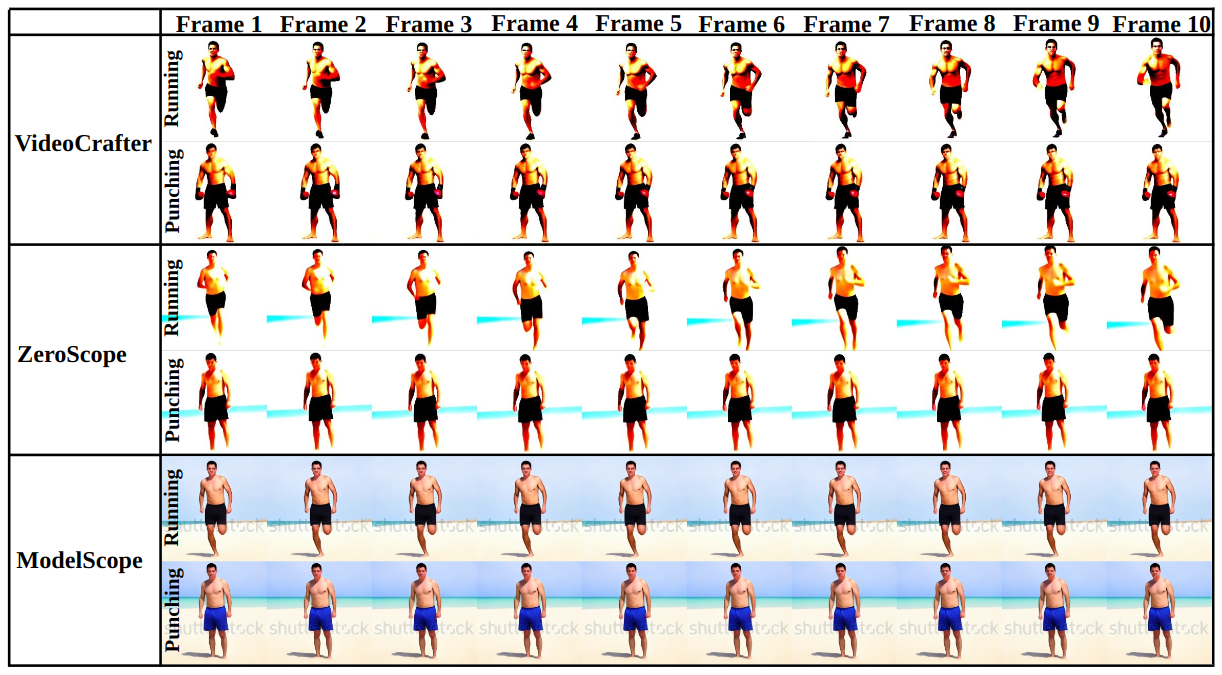}\vspace{-10pt}
    \caption{Visualization of the optimized latent when given the two actions as prompts. Generated videos in the top row of each video model denote the action "running". The bottom row of each denotes the action "punching". Clearly, the top rows of each model show a more natural motion.  VideoCrafter\cite{videocrafter2} demonstrates a higher degree of realism in both actions compared to other models}
    \label{fig:latent}
    \vspace{-7mm}
\end{figure}

\textbf{Evaluating MotionDistill:} Figure \ref{fig:results} showcases the capabilities of our method across various prompts. Our approach excels at generating plausible motion for common human activities, as demonstrated by the "walking" and "running" examples (Figure \ref{fig:results}(i) and \ref{fig:teaser}). However, challenges arise with less frequent actions.  The "punching" prompt (Figure \ref{fig:results}(ii)) resulted in partial motion generation, with limited hand movement. This shows for a semi-failure case. Furthermore, the model struggles with uncommon activities like "doing a cartwheel" (Figure \ref{fig:results}(iii)), failing to produce even a reasonable pose giving the extreme failure case 
\footnote{Check supplementary video at \href{https://github.com/Pauljanson002/human-eccv}{https://github.com/Pauljanson002/human-eccv} for sample clips}.\\
\textbf{Ablating MotionDistill:} It is challenging to determine if the failure cases comes from the SDS, the faithfulness of the video diffusion model, or the way we're representing motion and render it 
To gain deeper insights into the ability of SDS and our video models to produce faithful motion, we optimize our SDS objective with the
video diffusion model directly in their video latent space. 
We first render the initial video using Stage 1 of our pipeline.
Then detach this rendering from the optimization process and directly optimize the latents $\boldsymbol{Z}$. The results are shown in Figure \ref{fig:latent}.

Our experiments reveal performance discrepancies among three open-source video diffusion models, particularly concerning the generation of common versus rare human motions. This suggests a potential bias towards common activities in the training data. To illustrate this, we used two prompts for all three models: "running" (a common action) and "punching" (less frequent). As shown in Figure \ref{fig:latent}, the models generated more natural motion sequences for "running." In contrast, the "punching" frames lacked variation, highlighting the limitations of all three models with less common motions. We also observed that recent models\cite{videocrafter2} provide a much more realistic motion than older models\cite{modelscope,zeroscope}. This is evident, if we compare the top rows of each model, ModelScope\cite{modelscope} seems to have very little variation across frames even for "running". Indeed we observe when the model has challenges with the latent generation it also struggles in the case of fitting PoseField, suggesting the issue lies not in the representation of the body motion but in the video diffusion model.\\ 
\textbf{Implementation details:}
We implemented the pipeline in PyTorch\cite{pytorch} and used nvdiffrast\cite{nvdiffrast} for differentiable rendering. We conducted experiments in NVIDIA A6000 GPU. We used Adam\cite{adam} optimizer with a learning rate of 5e-4 for 10,000 iterations. Experiments maintained a CFG\cite{cfg} scale of 100. We initialized the PoseField to output the mean pose of the target motion and constrained its output within a range defined by three times the standard deviation of the target pose parameters. We set total frames $F$=10, $\lambda_{reg}$=1e-3 , $\lambda_{SDS_{t}}$=1e-3. 
\vspace{-5mm}
\section{Conclusion}
\vspace{-4mm}
Video diffusion models can generate human motion from text, but their performance differs a lot between familiar and rare actions. Notably, more recent models like VideoCrafter2\cite{videocrafter2} outperform earlier ones\cite{modelscope,zeroscope}. These findings underscore the need for further research to improve the diversity and quality of human motion generation, particularly for less frequent or complex actions. We hypothesize that our method when combined with more powerful text-to-video foundation models can become increasingly more effective.  Our work stands as proof of concept in this domain and studies the strengths and limitations of current open-source video diffusion models. Additionally, studying human motion understanding as an emergent behavior of a video diffusion model will be a promising future direction.
\clearpage
%
%
\bibliographystyle{splncs04}
\bibliography{main}

\end{document}